\pdfoutput=1

\documentclass[11pt]{article}

\usepackage{acl}

\usepackage{times}
\usepackage{latexsym}

\usepackage[T1]{fontenc}

\usepackage[utf8]{inputenc}

\usepackage{microtype}

\usepackage{inconsolata}

\usepackage{booktabs}
\usepackage{graphicx}
\usepackage{amsmath}
\usepackage{amssymb}
\usepackage{bbm}
\usepackage{arydshln}
\DeclareMathOperator*{\argmax}{arg\,max}

\usepackage[ruled,vlined,linesnumbered]{algorithm2e}
\usepackage{algpseudocode}
\usepackage{hyperref}
\usepackage{multirow}
\usepackage{colortbl}
\usepackage{bm}
\usepackage{soul}
\usepackage{adjustbox}
\usepackage{tcolorbox}
\usepackage{soul}
\tcbuselibrary{theorems,skins}
\usepackage{xcolor}
\definecolor{mygreen}{HTML}{CAE5CD}

\title{On the True Distribution Approximation of Minimum Bayes-Risk Decoding}

\author{
Atsumoto Ohashi$^{1}$\thanks{\ \ Work done during an internship at CyberAgent.}\qquad Ukyo Honda$^{2}$\qquad Tetsuro Morimura$^{2}$\qquad Yuu Jinnai$^{2}$\\
$^1$Nagoya University\qquad $^2$CyberAgent\\
\texttt{ohashi.atsumoto.c0@s.mail.nagoya-u.ac.jp}\\ \texttt{\{honda\_ukyo,morimura\_tetsuro,jinnai\_yu\}@cyberagent.co.jp}
}

\begin{document}
\maketitle
\begin{abstract}
Minimum Bayes-risk (MBR) decoding has recently gained renewed attention in text generation.
MBR decoding considers texts sampled from a model as pseudo-references and selects the text with the highest similarity to the others.
Therefore, sampling is one of the key elements of MBR decoding, and previous studies reported that the performance varies by sampling methods.
From a theoretical standpoint, this performance variation is likely tied to how closely the samples approximate the true distribution of references.
However, this approximation has not been the subject of in-depth study.
In this study, we propose using anomaly detection to measure the degree of approximation.
We first closely examine the performance variation and then show that previous hypotheses about samples do not correlate well with the variation, but our introduced anomaly scores do.
The results are the first to empirically support the link between the performance and the core assumption of MBR decoding.\footnote{The code is available at \url{https://github.com/CyberAgentAILab/mbr-anomaly}.}
\end{abstract}

\section{Introduction}
\label{sec:intro}
Minimum Bayes-risk (MBR) decoding has recently re-emerged as a better alternative to beam search in text generation such as neural machine translation (NMT), text summarization, and image captioning \citep{eikema-aziz-2020-map,freitag-etal-2022-high,fernandes-etal-2022-quality,suzgun-etal-2023-follow,bertsch2023its}.
MBR decoding first samples texts from a model and then selects the text most similar to the others, considering the text samples as substitutes for references.
Therefore, sampling plays an important role in MBR decoding, and previous studies have reported that the performance varies by sampling methods \citep{eikema-aziz-2020-map,eikema-aziz-2022-sampling,fernandes-etal-2022-quality,freitag2023epsilon}.

From a theoretical standpoint, the samples are assumed to approximate the \emph{true distribution}, the distribution of human-quality translations \citep{kumar-byrne-2002-minimum,kumar-byrne-2004-minimum}.
If the approximation deviates, biases can emerge in results of MBR decoding.
This implies a significant link between the performance variation and approximation quality.
Although previous studies explained the performance variation by some properties of samples, \emph{e.g.}, sampling bias and cumulative probability mass \citep{eikema-aziz-2020-map,freitag2023epsilon}, those properties have no clear relation with the true distribution.
Consequently, the relation between the performance gains by sampling methods and the core assumption remains unclear.

This study aims to empirically support the link between the performance and the approximation of true distribution.
To this end, we introduce measures for the degree of approximation.
If the assumption for samples holds, references, which are drawn from the true distribution by definition, should not deviate from the majority of the samples.
Based on this recasting, we employ \emph{anomaly detection} (also called \emph{outlier or novelty detection}) for the measure.
Our hypothesis is that references achieve lower anomaly scores among samples obtained with a higher-performance sampling method.
We first closely examine the performance variation by sampling methods.
Then, we show that the variation highly correlates with the anomaly scores but not so with the properties based on previous hypotheses.
The results are the first to provide empirical evidence for the link between the performance and core assumption, which is an important step to understanding the connection between the actual performance and the theory of MBR decoding.

\begin{figure}[t]
    \centering
    \includegraphics[width=1.0\columnwidth,keepaspectratio]{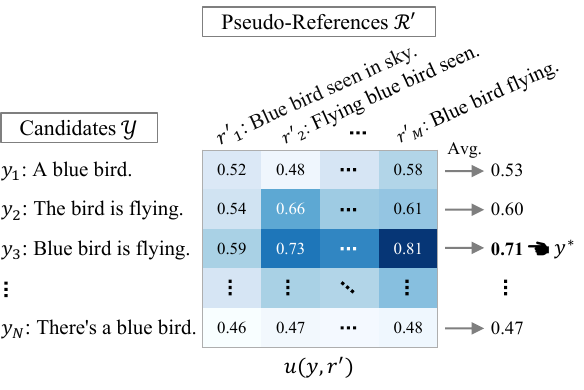}
    \caption{
    Illustrative example of MBR decoding.
    }
    \label{fig:mbr_overview}
\end{figure}

\section{Preliminaries}
\label{sec:preliminaries}
Let $u(y,r)$ be a utility function to measure the quality of model translation $y$ \citep[\textbf{candidate};][]{freitag-etal-2022-high} given its reference translation $r$.
Among a set of candidates $\mathcal{Y}$, MBR decoding selects the one that minimizes the expected error or, equivalently, maximizes the expected utility \citep{kumar-byrne-2002-minimum,kumar-byrne-2004-minimum,freitag-etal-2022-high}:
\begin{align}
    \label{eq:true mbr}
    y^* &= \argmax_{y \in \mathcal{Y}} \mathbb{E}_{r \sim P_{\mathrm{human}} (\cdot | x)} [u(y, r)].
\end{align}
Here, $P_{\mathrm{human}} (\cdot | x)$ is the \textbf{true distribution} over translations of an input text $x$ \citep{kumar-byrne-2002-minimum,kumar-byrne-2004-minimum}, which describes human-quality translations in the space of all translations.

Since the true distribution is unknown, MBR decoding approximates Eq.~\eqref{eq:true mbr} with finite samples drawn from a model $r' \sim P_{\mathrm{model}} (\cdot | x)$.
That is, MBR decoding \emph{assumes that the samples drawn from a model approximate the true distribution} of references.
The samples are called \textbf{pseudo-references} \citep{freitag-etal-2022-high}, which subsequently serve as alternatives to references in the computation of MBR as follows:
\vspace{-2pt}
\begin{align}
    \label{eq:approx mbr}
    y^* &= \argmax_{y \in \mathcal{Y}} \frac{1}{|\mathcal{R}'|} \sum_{r' \in \mathcal{R}'} u(y, r').
\end{align}
In practice, candidates $\mathcal{Y}$ and pseudo-references $\mathcal{R}'$ can be the same or different sets of samples.
Figure~\ref{fig:mbr_overview} shows an example of the above procedure.

\section{Performance Variation by Sampling}
\label{sec:sampling}
Previous studies reported that performance varies by sampling methods in NMT.
However, they used the same set of model translations for both candidates and pseudo-references \citep{eikema-aziz-2020-map,fernandes-etal-2022-quality,freitag2023epsilon} or explored sampling methods only for candidates \citep{eikema-aziz-2022-sampling}.
These settings obscure the effect of pseudo-references, for which the true distribution is assumed, on the performance variation.
This section shows the effect of pseudo-reference sampling on performance by evaluating pseudo-references separately from candidates.

\subsection{Setup}
\label{sec:setup sampling}
Following \citet{fernandes-etal-2022-quality}, we use publicly-available Transformer models \citep{vaswani2017attention} trained by \citet{ng-etal-2019-facebook}\footnote{\url{https://github.com/facebookresearch/fairseq/blob/7409af7f9a7b6ddac4cbfe7cafccc715b3c1b21e/examples/translation/README.md}} for the WMT19 news translation task \citep{barrault-etal-2019-findings}.
The models were trained in four directions between English (en) and German (de) or Russian (ru).
We conducted our experiments on the test set of WMT19 (\emph{newstest19}), which was used as the development set in the previous work \citep{fernandes-etal-2022-quality}.
Due to the quadratic computational cost of MBR decoding, we drew 100 samples of $\mathcal{Y}$ and $\mathcal{R}'$ for each of the 1,000 examples of \emph{newstest19}.
We employ COMET22 for the utility function $u$, which is the state-of-the-art evaluation metric in machine translation \citep{rei-etal-2022-comet,rei-etal-2020-comet}.\footnote{COMET22 improved robustness to the deviation in numbers and named entities, which was the weakness of the previous COMET \citep{amrhein-sennrich-2022-identifying}.}

For sampling methods, we use those that have been reported to be effective: ancestral sampling \citep{eikema-aziz-2020-map,freitag-etal-2022-high}, beam search, nucleus sampling \citep{eikema-aziz-2022-sampling,fernandes-etal-2022-quality}, and epsilon sampling \citep{freitag2023epsilon}.
Ancestral sampling draws samples from $P_{\mathrm{model}} (\cdot | x)$ without modification, while nucleus sampling restricts sampling to words with top-$p$ probabilities \citep{Holtzman2020The} and epsilon sampling truncates words with probabilities lower than $\epsilon$ \citep{hewitt-etal-2022-truncation}.
We adopt the best hyperparameters reported for $p$ and $\epsilon$ \citep{fernandes-etal-2022-quality,freitag2023epsilon}.
The beam size was set to 100 to collect 100 samples.

\subsection{Results}
\label{sec:results sampling}

\paragraph{Fixing Candidates.}
\label{sec:candidates selection}
Since we focus on sampling for pseudo-references, we first search for the best sampling method \emph{for candidates} and fix it.
The objective of searching for the best is to prevent the pseudo-reference’s contribution to scores from being capped and obscured by the candidate's quality.
To this end, we conduct the search on the same \emph{newstest19} as the subsequent experiments.\footnote{
If the objective is to find the best combination of sampling methods, which is not our focus, then it is desirable to use different splits to explore and test the combination to ensure the generalization.
Nevertheless, our subsequent results in Tables~\ref{tab:candidate}, \ref{tab:pseudo-reference}, \ref{tab:candidate-comet20}, and \ref{tab:pseudo-reference-comet20} suggest the generalization of the found best combination as it consistently performs the best across almost all language pairs.
}
Following \citet{fernandes-etal-2022-quality}, we search for the sampling method that achieves the highest oracle score, $\max_{y \in \mathcal{Y}} u(y,r)$, on average.
Table~\ref{tab:candidate} shows that epsilon sampling achieves the highest oracle score across the language pairs.
Based on these results, we fixed the sampling method of candidates to epsilon sampling in all the following experiments.

\paragraph{Varying Pseudo-References.}
\label{sec:pseudo-reference variation}
Then, we evaluate the effect of pseudo-references on performance by varying their sampling methods.
Table~\ref{tab:pseudo-reference} shows the results.
As expected from previous studies, the performance of MBR decoding varies even when only changing the sampling methods of pseudo-references.
The variation is nearly consistent across language pairs, indicating the pervasive effect of pseudo-reference on performance.
The best sampling method for candidates (epsilon sampling) is not the best for pseudo-references.
This shows that the desirable properties for candidates and pseudo-references are different.

Table~\ref{tab:pseudo-reference} also shows the results of beam search just for the comparison with MBR decoding.
Here, the beam size was set to 5.
MBR decoding significantly outperforms beam search and even outperforms the ensemble model, which was the winner of WMT19 \citep{barrault-etal-2019-findings}.
Since the effectiveness of epsilon sampling was reported on the other WMT dataset \citep{freitag2023epsilon}, we have a good reason to use epsilon sampling for this comparison.

\begin{table}[t]
\centering
\begin{adjustbox}{max width=\columnwidth}
\begin{tabular}{lcccc}
\toprule
\emph{Candidate} & de-en & en-de & ru-en & en-ru \\
\midrule
Ancestral & \underline{85.82} & \underline{86.32} & \underline{82.11} & \underline{86.13} \\
Beam & 88.47 & 89.32 & 84.16 & 89.44 \\
Epsilon ($\epsilon$ = 0.02) & \textbf{88.51} & \textbf{89.47} & \textbf{84.36} & \textbf{90.17} \\
Nucleus ($p$ = 0.6) & 88.01 & 89.12 & 83.76 & 89.96 \\
Nucleus ($p$ = 0.9) & 88.02 & 89.04 & 83.98 & 89.57 \\
\bottomrule
\end{tabular}
\end{adjustbox}
\caption{
\emph{Oracle scores} in COMET22 to determine the sampling method \emph{for candidates}.
The results are the average of three runs with different seeds except for beam search.
The best/worst scores are in \textbf{bold}/\underline{underlined}.
}
\label{tab:candidate}
\end{table}

\begin{table}[t]
\centering
\begin{adjustbox}{max width=\columnwidth}
\begin{tabular}{llcccc}
\toprule
\multirow{7}*{\rotatebox[origin=c]{90}{\emph{Epsilon ($\epsilon$ = 0.02)}}} & \emph{Pseudo-Reference} & de-en & en-de & ru-en & en-ru \\
\cmidrule(lr){2-6}
 & Ancestral & 85.82 & 87.51 & 82.02 & 88.41 \\
 & Beam & \underline{85.62} & \underline{87.40} & \underline{81.64} & \underline{87.78} \\
 & Epsilon ($\epsilon$ = 0.02) & 85.89 & 87.74 & 82.01 & 88.46 \\
 & Epsilon ($\epsilon$ = 0.02)$^*$ & 85.87 & 87.74 & 81.98 & 88.46 \\
 & Nucleus ($p$ = 0.6) & 85.69 & 87.57 & 81.76 & 88.26 \\
 & Nucleus ($p$ = 0.9) & \textbf{86.04} & \textbf{87.82} & \textbf{82.18} & \textbf{88.61} \\
\midrule
\midrule
 & Beam Search & 84.38 & 86.13 & 80.76 & 85.69 \\
 & Beam Search (ensemble) & 84.30 & 86.06 & 80.91 & 85.74 \\
\bottomrule
\end{tabular}
\end{adjustbox}
\caption{
COMET22 scores of MBR decoding with different pseudo-references.
Candidates are sampled with epsilon sampling ($\epsilon$ = 0.02).
Epsilon ($\epsilon$ = 0.02)$^*$ shows the results of sampling candidates and pseudo-references with the same epsilon sampling but with different seeds.
The results are the average of three runs with different seeds except for beam search.
The best/worst scores are in \textbf{bold}/\underline{underlined}.
}
\label{tab:pseudo-reference}
\end{table}

\section{Hypotheses for Performance Variation}
\label{sec:hypothesis}
The previous section confirmed that the performance varies by sampling pseudo-references.
The question that naturally arises in response to the results is: why does this variation occur?

\subsection{Previous Hypotheses}
\label{sec:previous hyp}
\citet{eikema-aziz-2022-sampling} hypothesized that unbiased sampling is desirable for pseudo-references.
Since the biased sampling methods limit the sampling to words of high probability, we use the average log probability (\textbf{Avg. Prob.}) of samples as a continuous proxy of bias existence in sampling. 
\citet{eikema-aziz-2020-map} and \citet{freitag2023epsilon} did not distinguish between candidates and pseudo-references but referred to the larger cumulative probability mass (\textbf{Cum. Prob.}) of unique samples as a desirable property because it indicates diverse and probable samples.
\citet{eikema-aziz-2022-sampling} employed candidate sampling that achieved high expected utility.
If this criterion applies to pseudo-references, performance should be higher when the expected utility against candidates (\textbf{Cand. Sim.}) or references (\textbf{Ref. Sim.}) is high.

\subsection{Our Hypothesis}
\label{sec:our hyp}
Given the relaxation from Eq.~\eqref{eq:true mbr} to Eq.~\eqref{eq:approx mbr}, a better approximation of the true distribution by pseudo-references should be associated with higher performance.
To examine the relation, we propose using anomaly detection to quantitatively evaluate the approximation.
If a better approximation is achieved, references should deviate less from the majority of the samples since references are drawn from the true distribution by definition.
This recasting allows us to use anomaly scores of anomaly detection for measuring the degree of approximation.
We then hypothesize that \emph{a higher-performance sampling method forms samples where references achieve lower anomaly scores}.

\section{Experiments}
\label{sec:experiments}
We test the hypotheses discussed in the previous section by evaluating the correlation between the performance variation and the properties or anomaly scores.

\subsection{Setup}
\label{sec:setup correlation}
The setup is the same as described in Section~\ref{sec:setup sampling}.
We run each sampling method with three different seeds and then calculate the Spearman's rank correlation coefficient $\rho$ between their averaged properties or anomaly scores (see Section~\ref{sec:hypothesis}) with the COMET22 scores reported in Table~\ref{tab:pseudo-reference}.

\subsection{Anomaly Scores}
\label{sec:measures}
To test our hypothesis, we employ three popular methods used in \emph{unsupervised} anomaly detection \citep{kriegel2011interpreting,gu2019statistical,ruff2021unifying}.
The first uses \textbf{Mahalanobis distance} \citep[$\bm{d}_{\bm{M}}$;][]{mahalanobis1936generalized} as an anomaly score, a classical distance measure between a data point and a distribution.
In our context, the distance is between reference $r$ and pseudo-references $\mathcal{R}'$ in a feature space: $ d_M(r, \mathcal{R}') = \sqrt{(r-\mu) \Sigma^{-1} (r-\mu)}$, where $\mu$ and $\Sigma^{-1}$ are mean and inverse covariance matrix of $\mathcal{R}'$, respectively.
Mahalanobis distance assumes that the data is normally distributed, but this assumption does not necessarily hold.
\textbf{$\bm{k}$-nearest neighbors} \citep[$\bm{k}\mathbf{NN}$;][]{Angiulli2002fast} does not have the assumption and is applicable to other data, such as the one with multimodal distribution.
$k\mathrm{NN}$ is a simple algorithm to consider the local density of a given data point.
Still, it is known to perform favorably to some state-of-the-art algorithms for anomaly detection \citep{gu2019statistical}.
Let $N_i(r, \mathcal{R}')$ be the $i$th nearest pseudo-reference to $r$ in Euclidean distance.
$k\mathrm{NN}$ takes the average of Euclidean distance from $r$ to its $k$-nearest pseudo-references $\{N_i(r, \mathcal{R}')\}_{i=1}^k$: $k\mathrm{NN}(r, \mathcal{R}') = \frac{1}{k}\sum_{i=1}^k \lVert r - N_i(r, \mathcal{R}') \rVert$.
\textbf{Local outlier factor} \citep[$\mathbf{LOF}$;][]{breunig2000lof} additionally considers the local density of the $k$-nearest data points themselves.
$\mathrm{LOF}$ in our setting measures how the local density of $r$ deviates from that of its $k$-nearest pseudo-references.
To better illustrate the relationship with $k\mathrm{NN}$, we show a simplified version of $\mathrm{LOF}$ \citep{schubert2014local} here: $\mathrm{LOF}_k(r, \mathcal{R}') = \frac{1}{k} \sum_{r' \in \{N_i(r, \mathcal{R}')\}_{i=1}^k} \frac{\lVert r - N_k(r, \mathcal{R}') \rVert}{\lVert r' - N_k(r', \mathcal{R}') \rVert}$.
\vspace{3pt}
See \citet{breunig2000lof} for the complete formula we used.

To calculate the anomaly scores, samples needs to be represented in a feature space.
We obtain the representation in the space of utility by measuring the utility of references or pseudo-references given a set of candidates $\mathcal{Y}$.
A reference $r$ in the space is then defined as $\begin{bmatrix} u(r, y_1), \dots, u(r, y_{|\mathcal{Y}|}) \end{bmatrix}^{\top}$.
Same for a pseudo-reference $r'$.

\begin{table}[t]
\centering
\begin{adjustbox}{max width=\columnwidth}
\begin{tabular}{lcccc}
\toprule
 & de-en & en-de & ru-en & en-ru \\
\midrule
Avg. Prob.$_{{(-)}}$ & 0.580$_{(\checkmark)}$ & 0.290$_{(\checkmark)}$ & 0.870$_{(\checkmark)}$ & 0.638$_{(\checkmark)}$ \\
Cum. Prob.$_{(+)}$ & \underline{0.058}$_{(\textcolor{red}{\textcolor{red}{\times}})}$ & \underline{0.116}$_{(\textcolor{red}{\times})}$ & \underline{0.348}$_{(\textcolor{red}{\times})}$ & \underline{0.058}$_{(\textcolor{red}{\times})}$ \\
Cand. Sim.$_{(+)}$ & 0.543$_{(\textcolor{red}{\times})}$ & 0.314$_{(\textcolor{red}{\times})}$ & 0.829$_{(\textcolor{red}{\times})}$ & 0.657$_{(\textcolor{red}{\times})}$ \\
Ref. Sim.$_{(+)}$ & 0.580$_{(\textcolor{red}{\times})}$ & 0.290$_{(\textcolor{red}{\times})}$ & 0.870$_{(\textcolor{red}{\times})}$ & 0.638$_{(\textcolor{red}{\times})}$ \\
\midrule
$d_{M{(-)}}$ & 0.771$_{(\checkmark)}$ & 0.486$_{(\checkmark)}$ & 0.886$_{(\checkmark)}$ & 0.771$_{(\checkmark)}$ \\
$k\mathrm{NN}_{(-)}$ &  &  &  &  \\
\ \ \ \ $k$ = 5 & 0.771$_{(\checkmark)}$ & 0.829$_{(\checkmark)}$ & 0.886$_{(\checkmark)}$ & 0.829$_{(\checkmark)}$ \\
\ \ \ \ $k$ = 25 & 0.943$_{(\checkmark)}$ & \textbf{0.943}$_{(\checkmark)}$ & 0.886$_{(\checkmark)}$ & \textbf{0.943}$_{(\checkmark)}$ \\
\ \ \ \ $k$ = 50 & 0.771$_{(\checkmark)}$ & \textbf{0.943}$_{(\checkmark)}$ & \textbf{0.943}$_{(\checkmark)}$ & 0.829$_{(\checkmark)}$ \\
\ \ \ \ $k$ = 75 & 0.771$_{(\checkmark)}$ & \textbf{0.943}$_{(\checkmark)}$ & 0.371$_{(\checkmark)}$ & 0.829$_{(\checkmark)}$ \\
\ \ \ \ $k$ = 100 & 0.086$_{(\checkmark)}$ & 0.314$_{(\checkmark)}$ & 0.371$_{(\checkmark)}$ & 0.029$_{(\checkmark)}$ \\
$\mathrm{LOF}_{(-)}$ &  &  &  &  \\
\ \ \ \ $k$ = 5 & 0.829$_{(\checkmark)}$ & 0.600$_{(\checkmark)}$ & 0.943$_{(\checkmark)}$ & 0.771$_{(\checkmark)}$ \\
\ \ \ \ $k$ = 25 & 0.829$_{(\checkmark)}$ & 0.714$_{(\checkmark)}$ & \textbf{0.943}$_{(\checkmark)}$ & 0.829$_{(\checkmark)}$ \\
\ \ \ \ $k$ = 50 & \textbf{1.000}$_{(\checkmark)}$ & 0.886$_{(\checkmark)}$ & \textbf{0.943}$_{(\checkmark)}$ & 0.829$_{(\checkmark)}$ \\
\ \ \ \ $k$ = 75 & \textbf{1.000}$_{(\checkmark)}$ & 0.886$_{(\checkmark)}$ & \textbf{0.943}$_{(\checkmark)}$ & 0.829$_{(\checkmark)}$ \\
\ \ \ \ $k$ = 100 & 0.600$_{(\checkmark)}$ & 0.371$_{(\checkmark)}$ & 0.886$_{(\checkmark)}$ & 0.657$_{(\checkmark)}$ \\
\bottomrule
\end{tabular}
\end{adjustbox}
\caption{
Correlation coefficients (Spearman's $\rho$) between COMET22 performance variation and pseudo-references' properties or anomaly scores.
We show the absolute value of $\rho$.
The subscript signs $_{(+/-)}$ are the expected signs of $\rho$ (see Section~\ref{sec:hypothesis}), and the subscript marks $_{(\checkmark/\textcolor{red}{\times})}$ show whether the actual signs match/mismatch the expected ones.
The best/worst scores are in \textbf{bold}/\underline{underlined}.
}
\label{tab:correlation}
\end{table}

\subsection{Results}
\label{sec:results}
Table~\ref{tab:correlation} shows the results.
As expected, the anomaly scores are clearly more correlated than the properties based on previous hypotheses.
Except for Cum. Prob., Cand. Sim., and Ref. Sim., the signs of $\rho$ are all as expected, including the anomaly scores.
See Table~\ref{tab:detail} in Appendix for the results used to calculate $\rho$.

Among the anomaly scores,\footnote{We took the median of $d_M$ and $\mathrm{LOF}$ scores instead of the mean because they are unstable due to the inverse covariant matrix $\Sigma^{-1}$ and division, respectively. For $d_M$, we removed duplicated $y$ from the position vector and added an identity matrix not to drop the rank of $\Sigma$ and stabilize the computation of $\Sigma^{-1}$. The value of the elements of the identity matrix was set to 1e-5, taking into account that the average value of the diagonal components of $\Sigma$ was 1e-3.} $k\mathrm{NN}$ and $\mathrm{LOF}$ with $k$ = 50 stably correlate with the performance variation better than those with $k$ = 100 and $d_M$.
We speculate that the significant degradation of $k$NN with $k$ = 100 is caused by outliers in pseudo-references.
While $k$NN with $k$ < 100 can effectively avoid including these outliers in the calculation of anomaly scores, $k$NN with $k$ = 100 cannot, and its anomaly scores are likely to be distorted by the outliers.
These results suggest that even if some pseudo-references are outliers against a reference, the performance tends to be higher if the rest of the pseudo-reference is close to the reference.
In other words, pseudo-references do not have to be close to references in entirety to perform well.

\section{Related Work}
\label{sec:related work}
MBR decoding has been used in automatic speech recognition \citep{goel2000mbr}, statistical machine translation \citep{kumar-byrne-2002-minimum,kumar-byrne-2004-minimum}, and NMT \citep{stahlberg-etal-2017-neural,shu2017later,blain-etal-2017-exploring}.
Recently, MBR decoding has gained prominence again in NMT because of the following two innovations.
(1)~\citet{eikema-aziz-2020-map} showed that MBR decoding with stochastic sampling has a potential to outperform MAP decoding methods, including beam search; (2)~\citet{freitag-etal-2022-high} and \citet{fernandes-etal-2022-quality} explored utility functions and found that using neural reference-based metrics as the utility function significantly enhances the quality of output texts.
\citet{muller-sennrich-2021-understanding} reported domain robustness and less hallucination in the outputs of MBR decoding.
Other text generation tasks such as text summarization, image captioning, and diversity-aware text generation also benefit from MBR decoding \citep{suzgun-etal-2023-follow,borgeaud-emerson-2020-leveraging,jinnai2024generating}.
Recent studies have focused on improving the efficiency of MBR decoding \citep{cheng-vlachos-2023-faster,finkelstein2023mbr,yang2023direct,jinnai2023model,jinnai2024hyperparameter}.

The most related studies explored sampling methods for MBR decoding and raised hypotheses to explain the difference in performance by sampling methods \citep{eikema-aziz-2020-map,eikema-aziz-2022-sampling,fernandes-etal-2022-quality,freitag2023epsilon}.
We also explored sampling methods but differed in that we did it more closely by focusing on pseudo-references.
Furthermore, we introduced anomaly scores that correlate with the performance variation better than previous hypotheses.

\section{Conclusion}
\label{sec:conclusion}
This study investigated the relation between the performance of MBR decoding and the core assumption about samples: samples follow the true distribution of references.
We introduced anomaly scores used in anomaly detection to evaluate the approximation of the true distribution.
Experimental results demonstrated that the anomaly scores correlate with the performance significantly better than the properties hypothesized to explain the performance variation in prior literature.
The previous hypotheses assumed that unbiased sampling (Avg. Prob.), diverse and probable samples (Cum. Prob.), or high expected utility (Cand. and Ref. Sim.) are the key properties of samples to achieve high performance.
However, these properties do not have an obvious relationship to approximating the true distribution of references, in contrast to the anomaly scores we employed.

These results show the insufficiency of existing hypotheses about the properties that samples should possess.
The results are also the first to empirically support the link between the actual performance and the key assumption of MBR decoding.
We believe this serves as an essential step to understanding the connection between the actual performance and the theory of MBR decoding.

\section{Limitations and Risks}
\label{sec:limitations}
The limitation of the study is that it is solely a thorough analysis of MBR decoding, not accompanied by an algorithm to improve the performance of MBR decoding.
However, our analysis empirically shows that previous hypotheses about the properties of samples are insufficient and that following the assumption of the MBR decoding is the key to improving performance.
We believe this is an important contribution that modifies the direction of future development of MBR decoding.

Our investigation is limited to Transformer models provided by \citet{ng-etal-2019-facebook} and the task is limited to machine translation.
Future work will extend the analysis to a wider range of models and text generation tasks.
However, it is worth noting that some studies support the general applicability of MBR decoding findings obtained in NMT to other text generation tasks and models.
Some hyperparameters \citep{suzgun-etal-2023-follow}, efficiency-boosting techniques \citep{jinnai2023model,jinnai2024hyperparameter}, or diversity-aware extensions \citep{jinnai2024generating} for MBR decoding consistently perform well across machine translation, summarization, image captioning, and data-to-text generation with different models.
\citet{bertsch2023its} shows that MBR decoding works well even in open-ended text generation tasks.

We do not foresee any ethical concerns in our analysis.

\bibliography{anthology,custom}

\appendix

\clearpage

\section{Results with COMET20}
\label{sec:appendix comet20}
To support the analysis of this study with a different utility function, we conducted the same experiments as in Tables~\ref{tab:candidate}, \ref{tab:pseudo-reference}, and \ref{tab:correlation} using COMET20 \citep{rei-etal-2020-comet}.
Tables~\ref{tab:candidate-comet20}, and \ref{tab:pseudo-reference-comet20}, \ref{tab:correlation-comet20} show the same tendency: the performance varies by sampling methods almost consistently, and the anomaly scores achieve the best correlations with the performance variation.
These results confirm the validity of our analysis even with other utility functions.

\begin{table}[t]
\centering
\begin{adjustbox}{max width=\columnwidth}
\begin{tabular}{lcccc}
\toprule
\emph{Candidate} & de-en & en-de & ru-en & en-ru \\
\midrule
Ancestral & \underline{62.31} & \underline{61.38} & \underline{49.83} & \underline{64.40} \\
Beam & 73.27 & 71.82 & 58.08 & 78.19 \\
Epsilon ($\epsilon$ = 0.02) & \textbf{73.65} & \textbf{72.17} & \textbf{59.46} & \textbf{80.86} \\
Nucleus ($p$ = 0.6) & 71.11 & 70.93 & 56.56 & 80.12 \\
Nucleus ($p$ = 0.9) & 71.87 & 70.84 & 58.01 & 78.33 \\
\bottomrule
\end{tabular}
\end{adjustbox}
\caption{
\emph{Oracle scores} in COMET20 to determine the sampling method \emph{for candidates}.
The results are the average of three runs with different seeds except for beam search.
The best/worst scores are in \textbf{bold}/\underline{underlined}.
}
\label{tab:candidate-comet20}
\end{table}

\begin{table}[t]
\centering
\begin{adjustbox}{max width=\columnwidth}
\begin{tabular}{llcccc}
\toprule
\multirow{7}*{\rotatebox[origin=c]{90}{\emph{Epsilon ($\epsilon$ = 0.02)}}} & \emph{Pseudo-Reference} & de-en & en-de & ru-en & en-ru \\
\cmidrule(lr){2-6}
 & Ancestral & 58.77 & 64.21 & \textbf{47.92} & 71.19 \\
 & Beam & \underline{57.24} & \underline{63.49} & 46.56 & \underline{68.26} \\
 & Epsilon ($\epsilon$ = 0.02) & 59.07 & 65.32 & 46.56 & 71.85 \\
 & Epsilon ($\epsilon$ = 0.02)$^*$ & 59.19 & 65.25 & 46.59 & 71.79 \\
 & Nucleus ($p$ = 0.6) & 57.82 & 64.31 & \underline{45.42} & 71.08 \\
 & Nucleus ($p$ = 0.9) & \textbf{59.65} & \textbf{65.44} & 47.77 & \textbf{72.33} \\
\bottomrule
\end{tabular}
\end{adjustbox}
\caption{
COMET20 scores of MBR decoding with different pseudo-references.
Candidates are sampled with epsilon sampling ($\epsilon$ = 0.02).
Epsilon ($\epsilon$ = 0.02)$^*$ shows the results of sampling candidates and pseudo-references with the same epsilon sampling but with different seeds.
The results are the average of three runs with different seeds except for beam search.
The best/worst scores are in \textbf{bold}/\underline{underlined}.
}
\label{tab:pseudo-reference-comet20}
\end{table}

\begin{table}[t]
\centering
\begin{adjustbox}{max width=\columnwidth}
\begin{tabular}{lcccc}
\toprule
 & de-en & en-de & ru-en & en-ru \\
\midrule
Avg. Prob.$_{{(-)}}$ & 0.580$_{(\checkmark)}$ & 0.290$_{(\checkmark)}$ & 0.928$_{(\checkmark)}$ & 0.638$_{(\checkmark)}$ \\
Cum. Prob.$_{{(+)}}$ & \underline{0.058}$_{(\textcolor{red}{\times})}$ & \underline{0.116}$_{(\textcolor{red}{\times})}$ & 0.406$_{(\textcolor{red}{\times})}$ & \underline{0.058}$_{(\textcolor{red}{\times})}$ \\
Cand. Sim.$_{{(+)}}$ & 0.600$_{(\textcolor{red}{\times})}$ & 0.257$_{(\textcolor{red}{\times})}$ & 0.943$_{(\textcolor{red}{\times})}$ & 0.600$_{(\textcolor{red}{\times})}$ \\
Ref. Sim.$_{{(+)}}$ & 0.580$_{(\textcolor{red}{\times})}$ & 0.290$_{(\textcolor{red}{\times})}$ & 0.928$_{(\textcolor{red}{\times})}$ & 0.638$_{(\textcolor{red}{\times})}$ \\
\midrule
$d_{M{(-)}}$ & 0.714$_{(\checkmark)}$ & 0.543$_{(\checkmark)}$ & 0.886$_{(\checkmark)}$ & 0.829$_{(\checkmark)}$ \\
$k\mathrm{NN}_{{(-)}}$ &  &  &  &  \\
 \ \ \ $k$ = 5 & 0.829$_{(\checkmark)}$ & 0.714$_{(\checkmark)}$ & \textbf{1.000}$_{(\checkmark)}$ & 0.771$_{(\checkmark)}$ \\
 \ \ \ $k$ = 25 & \textbf{1.000}$_{(\checkmark)}$ & \textbf{0.886}$_{(\checkmark)}$ & \textbf{1.000}$_{(\checkmark)}$ & \textbf{0.886}$_{(\checkmark)}$ \\
 \ \ \ $k$ = 50 & 0.829$_{(\checkmark)}$ & \textbf{0.886}$_{(\checkmark)}$ & 0.943$_{(\checkmark)}$ & 0.771$_{(\checkmark)}$ \\
 \ \ \ $k$ = 75 & 0.829$_{(\checkmark)}$ & \textbf{0.886}$_{(\checkmark)}$ & 0.143$_{(\checkmark)}$ & 0.771$_{(\checkmark)}$ \\
 \ \ \ $k$ = 100 & 0.143$_{(\checkmark)}$ & 0.257$_{(\checkmark)}$ & \underline{0.086}$_{(\checkmark)}$ & 0.086$_{(\checkmark)}$ \\
$\mathrm{LOF}_{{(-)}}$ &  &  &  &  \\
 \ \ \ $k$ = 5 & 0.771$_{(\checkmark)}$ & 0.829$_{(\checkmark)}$ & 0.886$_{(\checkmark)}$ & 0.829$_{(\checkmark)}$ \\
 \ \ \ $k$ = 25 & 0.771$_{(\checkmark)}$ & 0.829$_{(\checkmark)}$ & 0.943$_{(\checkmark)}$ & 0.829$_{(\checkmark)}$ \\
 \ \ \ $k$ = 50 & 0.771$_{(\checkmark)}$ & 0.829$_{(\checkmark)}$ & 0.943$_{(\checkmark)}$ & 0.829$_{(\checkmark)}$ \\
 \ \ \ $k$ = 75 & 0.829$_{(\checkmark)}$ & 0.829$_{(\checkmark)}$ & 0.943$_{(\checkmark)}$ & 0.829$_{(\checkmark)}$ \\
 \ \ \ $k$ = 100 & 0.657$_{(\checkmark)}$ & 0.486$_{(\checkmark)}$ & \textbf{1.000}$_{(\checkmark)}$ & 0.486$_{(\checkmark)}$ \\
\bottomrule
\end{tabular}
\end{adjustbox}
\caption{
Correlation coefficients (Spearman's $\rho$) between COMET20 performance variation and pseudo-references' properties or anomaly scores.
We show the absolute value of $\rho$.
The subscript signs $_{(+/-)}$ are the expected signs of $\rho$ (see Section~\ref{sec:hypothesis}), and the subscript marks $_{(\checkmark/\textcolor{red}{\times})}$ show whether the actual signs match/mismatch the expected ones.
The best/worst scores are in \textbf{bold}/\underline{underlined}.
}
\label{tab:correlation-comet20}
\end{table}

\section{Detailed Results}
\label{sec:appendix results}
Tables~\ref{tab:detail} and \ref{tab:detail-comet20} show the results used to calculate the Spearman's $\rho$ in Tables~\ref{tab:correlation} and \ref{tab:correlation-comet20}, respectively.

\begin{table*}[t]
\centering
\begin{adjustbox}{max width=\textwidth}
\begin{tabular}{llcccccccccccccccc}
\toprule
 &  &  & \multicolumn{2}{c}{Prob.} & \multicolumn{2}{c}{Sim.} &  & \multicolumn{5}{c}{$k\mathrm{NN}$$\downarrow$} & \multicolumn{5}{c}{$\mathrm{LOF}$$\downarrow$} \\ 
\cmidrule(lr){4-5}
\cmidrule(lr){6-7}
\cmidrule(lr){9-13}
\cmidrule(lr){14-18}
 & \emph{Pseudo-Reference} & COMET22$\uparrow$ & Avg.$\downarrow$ & Cum.$\uparrow$ & Cand.$\uparrow$ & Ref.$\uparrow$ & $d_M$$\downarrow$ & 5 & 25 & 50 & 75 & 100 & 5 & 25 & 50 & 75 & 100 \\
\midrule
\rowcolor{mygreen}
 & \multicolumn{17}{c}{\textbf{de-en}} \\
\multirow{6}*{\rotatebox[origin=c]{90}{\emph{Epsilon ($\epsilon$ = 0.02)}}} 
 & Ancestral & 85.82 & \textbf{-3.59} & 0.87 & \underline{71.20} & \underline{60.45} & 8.02 & 0.22 & 0.39 & \underline{0.62} & \underline{0.88} & \underline{1.30} & 1.05 & 1.07 & 1.10 & 1.03 & 1.00 \\
 & Beam & \underline{85.62} & -0.76 & \textbf{1.02} & 85.44 & 83.01 & 15.10 & \underline{0.33} & \underline{0.41} & 0.46 & 0.50 & 0.54 & \underline{1.75} & \underline{1.55} & \underline{1.47} & \underline{1.25} & 1.00 \\
 & Epsilon ($\epsilon$ = 0.02) & 85.89 & -0.89 & 0.97 & 84.81 & 82.18 & \textbf{6.94} & 0.26 & 0.35 & 0.41 & 0.46 & 0.53 & 1.08 & 1.09 & 1.08 & 1.01 & 1.00 \\
 & Epsilon ($\epsilon$ = 0.02)* & 85.87 & -0.89 & 0.97 & 84.70 & 82.18 & 8.61 & 0.23 & 0.33 & 0.39 & 0.45 & \textbf{0.51} & 1.10 & 1.10 & 1.09 & 1.02 & 1.00 \\
 & Nucleus ($p$ = 0.6) & 85.69 & \underline{-0.70} & \underline{0.83} & \textbf{85.63} & \textbf{83.24} & 16.47 & 0.32 & 0.39 & 0.45 & 0.49 & 0.53 & 1.62 & 1.43 & 1.37 & 1.16 & 1.00 \\
 & Nucleus ($p$ = 0.9) & \textbf{86.04} & -1.50 & 0.95 & 81.66 & 78.02 & 7.38 & \textbf{0.18} & \textbf{0.27} & \textbf{0.35} & \textbf{0.42} & 0.57 & \textbf{1.04} & \textbf{1.02} & \textbf{1.01} & \textbf{0.99} & 1.00 \\
\midrule
\rowcolor{mygreen}
 & \multicolumn{17}{c}{\textbf{en-de}} \\
\multirow{6}*{\rotatebox[origin=c]{90}{\emph{Epsilon ($\epsilon$ = 0.02)}}} 
 & Ancestral & 87.51 & \textbf{-3.60} & 0.65 & \underline{68.41} & \underline{51.56} & 8.16 & 0.22 & \underline{0.44} & \underline{0.74} & \underline{1.09} & \underline{1.71} & 1.08 & 1.11 & 1.11 & 1.03 & 1.00 \\
 & Beam & \underline{87.40} & -0.65 & \textbf{0.74} & 87.02 & 85.06 & \underline{15.86} & \underline{0.30} & 0.37 & 0.41 & 0.44 & 0.47 & \underline{2.05} & \underline{1.80} & \underline{1.64} & \underline{1.36} & 1.00 \\
 & Epsilon ($\epsilon$ = 0.02) & 87.74 & -0.80 & 0.71 & 86.33 & 83.96 & \textbf{6.52} & 0.23 & 0.31 & 0.36 & 0.40 & 0.45 & 1.09 & 1.11 & 1.09 & 1.01 & 1.00 \\
 & Epsilon ($\epsilon$ = 0.02)* & 87.74 & -0.80 & 0.71 & 86.24 & 83.96 & 8.22 & 0.21 & 0.29 & 0.35 & 0.39 & \textbf{0.44} & 1.11 & 1.13 & 1.11 & 1.02 & 1.00 \\
 & Nucleus ($p$ = 0.6) & 87.57 & \underline{-0.61} & \underline{0.62} & \textbf{87.12} & \textbf{85.10} & 14.95 & 0.29 & 0.35 & 0.39 & 0.43 & 0.46 & 1.68 & 1.49 & 1.42 & 1.21 & 1.00 \\
 & Nucleus ($p$ = 0.9) & \textbf{87.82} & -1.30 & 0.70 & 83.84 & 79.38 & 7.08 & \textbf{0.16} & \textbf{0.24} & \textbf{0.31} & \textbf{0.37} & 0.49 & \textbf{1.04} & \textbf{1.04} & \textbf{1.02} & \textbf{1.00} & 1.00 \\
\midrule
\rowcolor{mygreen}
 & \multicolumn{17}{c}{\textbf{ru-en}} \\
\multirow{6}*{\rotatebox[origin=c]{90}{\emph{Epsilon ($\epsilon$ = 0.02)}}} 
 & Ancestral & 88.41 & \textbf{-3.75} & \underline{0.60} & \underline{70.67} & \underline{59.20} & 8.13 & \textbf{0.20} & \textbf{0.35} & 0.56 & \underline{0.83} & \underline{1.25} & \textbf{1.04} & \textbf{1.03} & \textbf{1.03} & \textbf{1.00} & \textbf{1.00} \\
 & Beam & \underline{87.78} & -0.74 & \textbf{0.74} & 85.71 & 79.69 & 24.41 & \underline{0.60} & \underline{0.68} & \underline{0.73} & 0.77 & 0.81 & \underline{3.02} & \underline{2.65} & \underline{2.40} & \underline{1.99} & \underline{1.28} \\
 & Epsilon ($\epsilon$ = 0.02) & 88.46 & -0.89 & 0.69 & 85.18 & 79.03 & 11.26 & 0.48 & 0.60 & 0.67 & 0.72 & 0.77 & 1.30 & 1.40 & 1.44 & 1.34 & \textbf{1.00} \\
 & Epsilon ($\epsilon$ = 0.02)* & 88.46 & -0.89 & 0.69 & 85.09 & 79.03 & 11.19 & 0.46 & 0.58 & 0.65 & 0.71 & 0.76 & 1.33 & 1.46 & 1.51 & 1.39 & \textbf{1.00} \\
 & Nucleus ($p$ = 0.6) & 88.26 & \underline{-0.69} & \underline{0.60} & \textbf{85.96} & \textbf{79.91} & \underline{25.16} & 0.59 & 0.67 & \underline{0.73} & 0.77 & 0.81 & 2.89 & 2.40 & 2.18 & 1.83 & 1.15 \\
 & Nucleus ($p$ = 0.9) & \textbf{88.61} & -1.52 & 0.67 & 82.07 & 75.17 & \textbf{7.85} & 0.24 & 0.41 & \textbf{0.52} & \textbf{0.61} & \textbf{0.75} & 1.05 & 1.09 & 1.14 & 1.02 & \textbf{1.00} \\
\midrule
\rowcolor{mygreen}
 & \multicolumn{17}{c}{\textbf{en-ru}} \\
\multirow{6}*{\rotatebox[origin=c]{90}{\emph{Epsilon ($\epsilon$ = 0.02)}}} 
 & Ancestral & 82.02 & \textbf{-3.85} & \underline{0.37} & \underline{71.25} & \underline{55.20} & 8.91 & 0.23 & 0.42 & \underline{0.67} & \underline{0.97} & \underline{1.40} & 1.07 & 1.13 & 1.13 & 1.05 & \textbf{1.00} \\
 & Beam & \underline{81.64} & \underline{-0.72} & \textbf{0.44} & \textbf{87.47} & \textbf{84.73} & \underline{20.91} & \underline{0.38} & \underline{0.44} & 0.48 & 0.51 & 0.54 & \underline{2.42} & \underline{2.04} & \underline{1.85} & \underline{1.51} & \underline{1.02} \\
 & Epsilon ($\epsilon$ = 0.02) & 82.01 & -0.94 & 0.42 & 86.68 & 83.57 & \textbf{7.27} & 0.28 & 0.37 & 0.42 & 0.46 & 0.51 & 1.10 & 1.12 & 1.11 & 1.03 & \textbf{1.00} \\
 & Epsilon ($\epsilon$ = 0.02)* & 81.98 & -0.94 & 0.42 & 86.58 & 83.57 & 9.04 & 0.26 & 0.35 & 0.40 & 0.45 & \textbf{0.50} & 1.11 & 1.14 & 114 & 1.05 & \textbf{1.00} \\
 & Nucleus ($p$ = 0.6) & 81.76 & -0.80 & \underline{0.37} & 87.06 & 84.26 & 13.17 & 0.30 & 0.38 & 0.43 & 0.47 & 0.51 & 1.35 & 1.36 & 1.35 & 1.17 & \textbf{1.00} \\
 & Nucleus ($p$ = 0.9) & \textbf{82.18} & -1.69 & 0.40 & 83.10 & 77.69 & 7.76 & \textbf{0.18} & \textbf{0.26} & \textbf{0.33} & \textbf{0.40} & 0.55 & \textbf{1.03} & \textbf{1.02} & \textbf{1.01} & \textbf{0.99} & \textbf{1.00} \\
\bottomrule
\end{tabular}
\end{adjustbox}
\caption{
Results used to calculate the Spearman's $\rho$ in Table~\ref{tab:correlation}.
Candidates are sampled with epsilon sampling ($\epsilon$ = 0.02).
Epsilon ($\epsilon$ = 0.02)$^*$ shows the results of sampling candidates and pseudo-references with the same epsilon sampling but with different seeds.
Avg. Prob. is the log probability.
$\uparrow$ and $\downarrow$ denote that the values are considered to be better when they are higher and lower, respectively.
The best/worst scores are in \textbf{bold}/\underline{underlined}.
}
\label{tab:detail}
\end{table*}

\begin{table*}[t]
\centering
\begin{adjustbox}{max width=\textwidth}
\begin{tabular}{llcccccccccccccccc}
\toprule
 &  &  & \multicolumn{2}{c}{Prob.} & \multicolumn{2}{c}{Sim.} &  & \multicolumn{5}{c}{$k\mathrm{NN}$$\downarrow$} & \multicolumn{5}{c}{$\mathrm{LOF}$$\downarrow$} \\ 
\cmidrule(lr){4-5}
\cmidrule(lr){6-7}
\cmidrule(lr){9-13}
\cmidrule(lr){14-18}
 & \emph{Pseudo-Reference} & COMET20$\uparrow$ & Avg.$\uparrow$ & Cum.$\uparrow$ & Cand.$\uparrow$ & Ref.$\uparrow$ & $d_M$$\downarrow$ & 5 & 25 & 50 & 75 & 100 & 5 & 25 & 50 & 75 & 100 \\
\midrule
\rowcolor{mygreen}
 & \multicolumn{17}{c}{\textbf{de-en}} \\
\multirow{6}*{\rotatebox[origin=c]{90}{\emph{Epsilon ($\epsilon$ = 0.02)}}} 
 & Ancestral                    & 58.77 & \underline{-3.59} & 0.87 & \underline{-0.11} & \underline{-0.51} & 17.65 & 1.13 & 1.98 & \underline{3.05} & \underline{4.41} & \underline{6.56} & \textbf{1.05} & 1.05 & 1.07 & 1.01 & 1.00 \\
 & Beam                         & \underline{57.24} & -0.76 & \textbf{1.02} & 0.59 & 0.46 & 46.37 & \underline{1.67} & \underline{2.08} & 2.34 & 2.55 & 2.81 & \underline{1.82} & \underline{1.62} & \underline{1.51} & \underline{1.28} & 1.00 \\
 & Epsilon ($\epsilon$ = 0.02)  & 59.07 & -0.89 & 0.97 & 0.56 & 0.43 & 21.33 & 1.27 & 1.76 & 2.10 & 2.37 & 2.69 & 1.12 & 1.13 & 1.11 & 1.01 & 1.00 \\
 & Epsilon ($\epsilon$ = 0.02)* & 59.19 & -0.89 & 0.97 & 0.56 & 0.43 & 23.26 & 1.19 & 1.69 & 2.03 & 2.31 & \textbf{2.64} & 1.13 & 1.14 & 1.11 & 1.02 & 1.00 \\
 & Nucleus ($p$ = 0.6)          & 57.82 & \textbf{-0.70} & \underline{0.83} & \textbf{0.60} & \textbf{0.48} & \underline{55.60} & 1.59 & 1.98 & 2.26 & 2.48 & 2.71 & 1.71 & 1.52 & 1.44 & 1.18 & 1.00 \\
 & Nucleus ($p$ = 0.9)          & \textbf{59.65} & -1.50 & 0.95 & 0.41 & 0.26 & \textbf{17.29} & \textbf{0.93} & \textbf{1.39} & \textbf{1.79} & \textbf{2.19} & 2.95 & \textbf{1.05} & \textbf{1.03} & \textbf{1.01} & \textbf{0.99} & 1.00 \\
\midrule
\rowcolor{mygreen}
 & \multicolumn{17}{c}{\textbf{en-de}} \\
\multirow{6}*{\rotatebox[origin=c]{90}{\emph{Epsilon ($\epsilon$ = 0.02)}}} 
 & Ancestral                    & 64.21 & \underline{-3.60} & 0.65 & \underline{-0.14} & \underline{-0.80} & 14.50 & 0.87 & \underline{1.62} & \underline{2.65} & \underline{4.15} & \underline{7.24} & 1.06 & 1.07 & 1.05 & \textbf{1.00} & 1.00 \\
 & Beam                         & \underline{63.49} & -0.65 & \textbf{0.75} & 0.62 & \textbf{0.56} & 40.90 & \underline{1.15} & 1.38 & 1.54 & 1.66 & 1.82 & \underline{2.06} & \underline{1.75} & \underline{1.57} & \underline{1.29} & 1.00 \\
 & Epsilon ($\epsilon$ = 0.02)  & 65.32 & -0.80 & 0.71 & 0.60 & 0.52 & 16.66 & 0.90 & 1.19 & 1.40 & 1.57 & 1.78 & 1.10 & 1.10 & 1.07 & 1.01 & 1.00 \\
 & Epsilon ($\epsilon$ = 0.02)* & 65.25 & -0.80 & 0.71 & 0.60 & 0.52 & 18.18 & 0.84 & 1.14 & 1.36 & 1.52 & \textbf{1.73} & 1.13 & 1.13 & 1.09 & 1.01 & 1.00 \\
 & Nucleus ($p$ = 0.6)          & 64.31 & \textbf{-0.61} & \underline{0.62} & \textbf{0.63} & \textbf{0.56} & \underline{42.62} & 1.11 & 1.34 & 1.51 & 1.65 & 1.79 & 1.84 & 1.47 & 1.34 & 1.14 & 1.00 \\
 & Nucleus ($p$ = 0.9)          & \textbf{65.44} & -1.30 & 0.70 & 0.51 & 0.35 & \textbf{14.21} & \textbf{0.67} & \textbf{0.99} & \textbf{1.24} & \textbf{1.48} & 1.97 & \textbf{1.05} & \textbf{1.05} & \textbf{1.02} & \textbf{1.00} & 1.00 \\
\midrule
\rowcolor{mygreen}
 & \multicolumn{17}{c}{\textbf{ru-en}} \\
\multirow{6}*{\rotatebox[origin=c]{90}{\emph{Epsilon ($\epsilon$ = 0.02)}}} 
 & Ancestral                    & \textbf{47.92} & \underline{-3.75} & \underline{0.60} & \underline{-0.05} & \underline{-0.45} & \textbf{17.37} & \textbf{0.96} & \textbf{1.63} & 2.49 & \underline{3.65} & \underline{5.68} & \textbf{1.04} & \textbf{1.03} & \textbf{1.02} & \textbf{1.00} & \textbf{1.00} \\
 & Beam                         & \underline{44.90} & -0.74 & \textbf{0.74} & 0.62 & 0.36 & 68.23 & \underline{2.56} & \underline{2.94} & \underline{3.18} & 3.36 & 3.55 & 3.01 & 1.54 & \underline{2.40} & \underline{1.98} & \underline{1.15} \\
 & Epsilon ($\epsilon$ = 0.02)  & 46.56 & -0.89 & 0.69 & 0.59 & 0.34 & 29.48 & 1.99 & 2.54 & 2.88 & 3.13 & 3.38 & 1.31 & 1.44 & 1.45 & 1.28 & \textbf{1.00} \\
 & Epsilon ($\epsilon$ = 0.02)* & 46.59 & -0.89 & 0.69 & 0.59 & 0.34 & 30.97 & 1.89 & 2.47 & 2.82 & 3.07 & \textbf{3.33} & 1.33 & 1.47 & 1.49 & 1.31 & \textbf{1.00} \\
 & Nucleus ($p$ = 0.6)          & 45.42 & \textbf{-0.69} & \underline{0.60} & \textbf{0.63} & \textbf{0.37} & \underline{87.59} & 2.49 & 2.88 & 3.15 & 3.35 & 3.53 & \underline{3.12} & \underline{2.40} & 2.18 & 1.76 & 1.08 \\
 & Nucleus ($p$ = 0.9)          & 47.77 & -1.52 & 0.67 & 0.45 & 0.18 & 19.08 & 1.10 & 1.79 & \textbf{2.27} & \textbf{2.67} & \textbf{3.33} & 1.06 & 1.11 & 1.13 & 1.01 & \textbf{1.00} \\
\midrule
\rowcolor{mygreen}
 & \multicolumn{17}{c}{\textbf{en-ru}} \\
\multirow{6}*{\rotatebox[origin=c]{90}{\emph{Epsilon ($\epsilon$ = 0.02)}}} 
 & Ancestral                    & 71.19 & \underline{-3.85} & \underline{0.37} & \underline{0.02} & \underline{-0.57} & 16.23 & 1.03 & 1.71 & \underline{2.60} & \underline{3.81} & \underline{5.85} & 1.06 & 1.08 & 1.08 & 1.02 & 1.00 \\
 & Beam                         & \underline{68.26} & \textbf{-0.72} & \textbf{0.44} & \textbf{0.69} & \textbf{0.57} & \underline{58.19} & \underline{1.64} & \underline{1.96} & 2.16 & 2.32 & 2.50 & \underline{2.33} & \underline{1.91} & \underline{1.73} & \underline{1.41} & 1.00 \\
 & Epsilon ($\epsilon$ = 0.02)  & 71.85 & -0.94 & 0.42 & 0.65 & 0.52 & 16.62 & 1.21 & 1.61 & 1.88 & 2.10 & 2.36 & 1.09 & 1.12 & 1.10 & 1.02 & 1.00 \\
 & Epsilon ($\epsilon$ = 0.02)* & 71.79 & -0.94 & 0.42 & 0.65 & 0.52 & 20.76 & 1.12 & 1.53 & 1.80 & 2.03 & \textbf{2.28} & 1.11 & 1.13 & 1.12 & 1.03 & 1.00 \\
 & Nucleus ($p$ = 0.6)          & 71.08 & -0.80 & \underline{0.37} & 0.67 & 0.56 & 36.80 & 1.30 & 1.67 & 1.92 & 2.12 & 2.34 & 1.37 & 1.32 & 1.32 & 1.14 & 1.00 \\
 & Nucleus ($p$ = 0.9)          & \textbf{72.33} & -1.69 & 0.41 & 0.49 & 0.30 & \textbf{15.40} & \textbf{0.83} & \textbf{1.21} & \textbf{1.52} & \textbf{1.84} & 2.52 & \textbf{1.03} & \textbf{1.02} & \textbf{1.01} & \textbf{0.99} & 1.00 \\
\bottomrule
\end{tabular}
\end{adjustbox}
\caption{
Results used to calculate the Spearman's $\rho$ in Table~\ref{tab:correlation-comet20}.
Candidates are sampled with epsilon sampling ($\epsilon$ = 0.02).
Epsilon ($\epsilon$ = 0.02)$^*$ shows the results of sampling candidates and pseudo-references with the same epsilon sampling but with different seeds.
Avg. Prob. is the log probability.
$\uparrow$ and $\downarrow$ denote that the values are considered to be better when they are higher and lower, respectively.
The best/worst scores are in \textbf{bold}/\underline{underlined}.
}
\label{tab:detail-comet20}
\end{table*}

\end{document}